%% file: example.tex
\documentclass{article}
\usepackage[T1]{fontenc}

\usepackage[preprint]{corl_2026} 
\usepackage{amsmath}
\usepackage{amssymb}
\usepackage{mathtools}
\usepackage{amsthm}
\usepackage{booktabs}
\usepackage{caption}
 \usepackage{multirow}
\usepackage{tabularx}
\usepackage{bm}
\usepackage{graphicx}
\title{Unifying Policy Learning and State Prediction through Spatial Language Modeling}

\author{
  Minye Wu \qquad Zehao Wang \qquad Tinne Tuytelaars\\
  {\normalfont\itshape KU Leuven}
}

\hypersetup{
  pdftitle={Unifying Policy Learning and State Prediction through Spatial Language Modeling},
  pdfauthor={Minye Wu, Zehao Wang, Tinne Tuytelaars}
}

\begin{document}
\maketitle

\input{sec/0_abstract}

\begin{figure}[h]
    \centering
    \includegraphics[width=\linewidth]{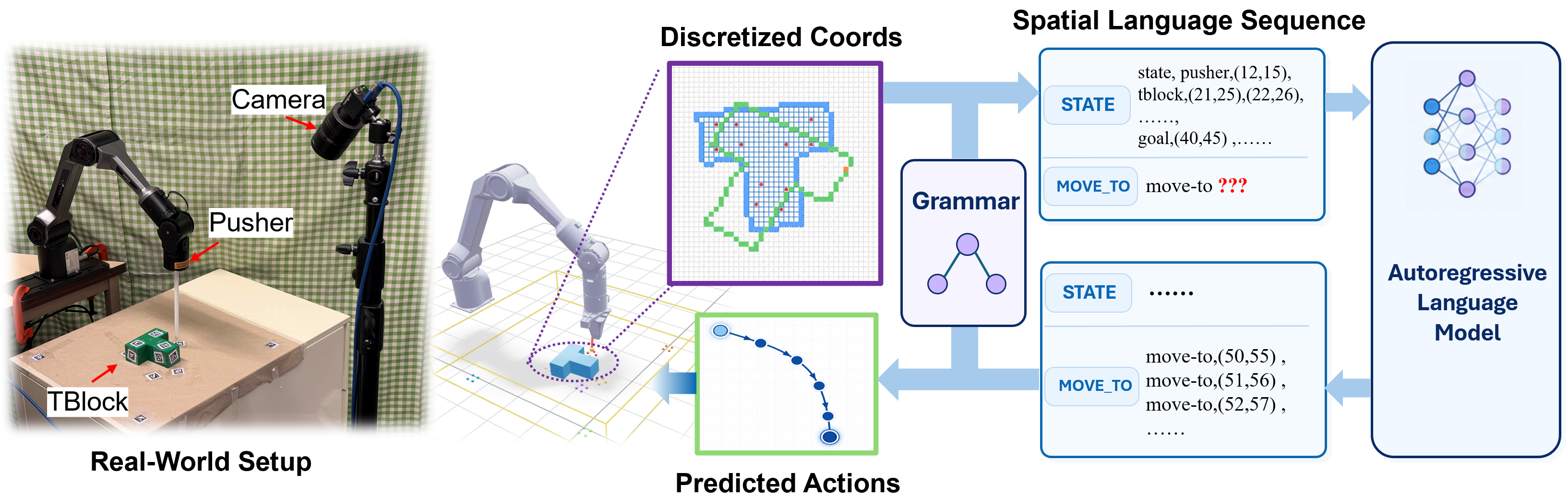}
    \caption{Spatial Language Modeling for Push-T. Scene geometry, goals, pusher actions, and optional future states are represented as structured coordinate-token sequences. The model generates actions for closed-loop execution and learns state transitions through future-state supervision.}
    \label{fig:teaser}
    \vspace{-3mm}
\end{figure}

\input{sec/1_intro}
\input{sec/related_work}
\input{sec/methods}
\input{sec/experiment}

\input{sec/Conclusion}


\clearpage

\bibliography{main}  

\clearpage
\appendix
\input{sec/supplementary}

\end{document}

%% file: sec/0_abstract.tex
\begin{abstract}
Learning how actions change scene geometry can provide complementary supervision for goal-directed manipulation. We introduce Spatial Language Modeling, which represents scene contours, goals, action targets, and future states with a shared vocabulary of discrete coordinates and semantic tokens. A task-specific grammar organizes these elements into spatial sequences, allowing one autoregressive Transformer to learn action generation and action-conditioned state prediction through a common next-token objective. We train the model from scratch using random-play transition pretraining followed by joint action and state training on expert demonstrations. During pretraining, recorded action coordinates condition subsequent state predictions and are excluded from the prediction loss. During control, the model decodes only executable action targets and updates its history with newly observed states. We evaluate the approach on Push-T in simulation and on a real robot. The model achieves competitive simulation performance and higher task success and target coverage than the evaluated real-robot policy baselines. Training ablations show improved control with joint action and state sequences, with further gains from random-play pretraining. Given supplied action trajectories, the same model also predicts successive scene states, capturing the geometric effects of pushing.
\end{abstract}

%% file: sec/1_intro.tex
\section{Introduction}
\label{sec:intro}

Robot manipulation requires selecting actions that transform a scene toward a goal. In the Push-T task~\cite{chi2025diffusion}, a planar pusher moves a T-shaped block to a target pose. The resulting motion depends on both object geometry and contact: the same pusher displacement can induce different combinations of translation and rotation depending on the contact location and the block's orientation. Demonstrations of these interactions provide two complementary training signals: expert actions indicate how to act, while subsequent observations reveal the geometric changes that follow.

Policies such as Action Chunking Transformers, Diffusion Policy, and vision-language-action models learn actions from demonstrations~\cite{zhao2023learning,chi2025diffusion,physicalintelligence2025pi05,bjorck2025gr00t}. World models use future predictions for rollout generation and planning~\cite{jang2025dreamgen, maes2026leworldmodel}, while recent work also jointly models actions and future observations~\cite{li2026causal}. Predicting the geometric consequences of actions can provide complementary supervision for policy learning and enable the use of non-expert interaction data. A shared prediction interface allows this transition supervision to update the same model parameters used for action generation. We therefore seek a spatial representation that makes both targets accessible to the same autoregressive model.

Discrete workspace coordinates provide such an interface at the level of explicit geometry. We divide a bounded planar workspace into grid cells and assign each cell a coordinate token. Object contours, target regions, pusher positions, and action trajectories can then be represented in the same spatial vocabulary. A coordinate retains the same physical reference whether it appears in an observed contour, an executable action target, or a predicted future state. Semantic tokens identify whether coordinates belong to the pusher, the manipulated object, the goal, or a movement command. Geometry and actions thus retain their distinct roles within a shared spatial reference frame.

We develop this representation as \textbf{Spatial Language Modeling}. A task-specific grammar organizes scene observations, goals, action targets, and optional subsequent states into a \emph{Spatial Language Sentence}(SLS). The grammar specifies how these elements are ordered and identified, while the model learns their coordinate content. A serializer converts observed geometry and recorded actions into spatial sentences, and a parser maps generated action blocks back to pusher targets. This formulation turns action generation and action-conditioned state prediction into sequence-completion tasks that share an output vocabulary and a next-token training objective.

We train an autoregressive Transformer from scratch using two sources of interaction data. Random-play pretraining first trains the model to predict scene transitions conditioned on recorded pusher motions. The action coordinates remain in the input sequence, but their prediction losses are masked; subsequent states supply targets describing the resulting geometry. Expert-demonstration training then supervises both action targets and selected subsequent states. These losses update the same parameters, making the geometric effects of interaction an explicit learning target alongside goal-directed action generation. This training scheme connects transition information from random interactions with action supervision from expert demonstrations.

During closed-loop control, an observed-history SLS prefix describes recent scene states and pusher actions. The model completes this prefix with consecutive action blocks, which are parsed into target coordinates and executed by the robot. Future-state blocks are excluded from control decoding, and new observations refresh the history for the next completion. The policy can therefore use a model trained with transition supervision while generating only actions at execution time. The same model also supports a separate prediction setting: given an initial scene and a supplied action trajectory, it generates successive scene states, using its previous state predictions as context.

We evaluate the formulation on Push-T in simulation and on a real robot. The model achieves competitive simulation performance and higher task success and target coverage than the evaluated real-robot policy baselines. Comparisons of training variants show improved control with interleaved action and state sequences, with further gains from random-play pretraining. In the state-prediction setting, predicted object states retain substantial geometric overlap with simulator reference states across the evaluated rollout horizon.

Our contributions are as follows:
\vspace{-2mm}
\begin{itemize}
\item We introduce a grammar-structured spatial representation that expresses observed geometry, goals, action targets, and future states in a shared coordinate vocabulary, enabling action and state prediction through autoregressive sequence completion.
\item We develop a training scheme that combines action-conditioned transition pretraining on random interactions with joint action and state supervision on expert demonstrations, and supports closed-loop execution through action-only decoding.
\item We demonstrate closed-loop Push-T control in simulation and on a real robot, and report  results and comparisons that support the proposed formulation.
\end{itemize}

%% file: sec/related_work.tex
\section{Related Work}

\vspace{2mm}
\noindent
\textbf{Policy Learning and World Modeling.}
Imitation-learning policies use demonstrations to supervise action distributions. Action Chunking Transformers predict temporally extended action sequences, while Diffusion Policy generates actions through trajectory denoising~\cite{zhao2023learning,chi2025diffusion}. Vision-language-action models connect action learning to vision-language pretraining, heterogeneous robot data, and fine-tuning~\cite{physicalintelligence2025pi05,kim2025fine,bjorck2025gr00t}. Sequence modeling also provides a direct route to control: Decision Transformer predicts actions conditioned on returns and trajectory history, and Behavior Transformers combine action discretization with continuous corrections to model multimodal demonstrations~\cite{chen2021decisiontransformer,shafiullah2022behavior}.

World models learn scene transitions and connect future predictions to control through several interfaces. DreamGen generates visual trajectories and obtains action labels through inverse dynamics or latent-action models for policy training~\cite{jang2025dreamgen}. LingBot-VA interleaves video prediction and action decoding in a causal sequence~\cite{li2026causal}. RoboDreamer synthesizes compositional video plans~\cite{zhou2024robodreamer}, Dreamitate tracks generated tool trajectories for robot execution~\cite{liang2024dreamitate}, and LeWorldModel predicts future latent embeddings for planning~\cite{maes2026leworldmodel}. Our formulation uses coordinate-level scene transitions and action targets within one next-token objective, combining random-play transition pretraining with joint action/state training on demonstrations.

\vspace{2mm}
\noindent
\textbf{Language Models in Spatial Computation.}
Language models have been applied to spatial intelligence through several representations of 3D scenes. PointLLM and 3D-LLM connect point-cloud or scene features to language models for object and scene understanding~\cite{xu2024pointllm,hong20233d}. SegPoint extends this connection to instruction-conditioned segmentation~\cite{he2024segpoint}. LSceneLLM adaptively selects task-relevant scene details, while Scene-LLM combines scene-level and egocentric features for 3D visual reasoning~\cite{zhi2025lscenellm,fu2025scene}. Other approaches expose spatial structure through prompting: SayPlan uses hierarchical 3D scene graphs for robot task planning~\cite{rana2023sayplan}, and 3DAxisPrompt supplies visual coordinate axes and masks for 3D grounding and reasoning~\cite{liu20253daxisprompt}.

Geometric sequence models provide another route to spatial representation. PolyGen autoregressively generates mesh vertices and faces, and MeshGPT generates discrete codes from a learned geometric vocabulary~\cite{nash2020polygen,siddiqui2024meshgpt}. Geometry-aware patch tokenization also supports mesh classification and segmentation~\cite{farazi2025recipe}. LLaMA-Mesh and MeshLLM connect mesh generation to language models through textual representations of geometry~\cite{wang2024llama,fang2025meshllm}. In 2D vision, Pix2Seq represents bounding boxes and class labels as discrete output tokens, formulating detection as sequence prediction~\cite{chen2021pix2seq}.

\vspace{2mm}
\noindent
\textbf{Scene Understanding and Generation.}
3D scene understanding develops geometric representations for identifying objects and their relationships. PointNet, PointNet++, PointConv, and Point Transformer learn features directly from point sets~\cite{qi2017pointnet,qi2017pointnet++,wu2019pointconv,zhao2021point}. VoxFormer predicts volumetric geometry and semantics from images~\cite{li2023voxformer}, while ScanRefer and context-aware grounding models connect language to objects in reconstructed scenes~\cite{chen2020scanrefer,yang2023exploiting}. Semantic scene graphs explicitly represent object relations~\cite{wald2020learning}, and PointCLIP models extend recognition using vision-language priors~\cite{zhang2022pointclip,zhu2023pointclip}. Complementary generative methods synthesize voxel shapes~\cite{wu2016learning}, surfaces and textured meshes~\cite{groueix2018papier,gao2022get3d}, or point clouds~\cite{yang2019pointflow}. DreamFusion, Magic3D, and Fantasia3D use text-to-image diffusion priors for text-conditioned 3D content creation~\cite{poole2022dreamfusion,lin2023magic3d,chen2023fantasia3d}. GALA3D organizes scene generation through layout-guided Gaussian representations~\cite{zhou2402gala3d}, while SmartSpatial uses depth and attention control to improve spatial arrangements in generated images~\cite{huang2025smartspatial}.

Across these directions, spatial representations provide the interface between scene geometry and task-dependent predictions. Our work organizes object contours, goals, action targets, and future-state targets into coordinate and semantic token sequences, with their roles and temporal order specified by a task grammar. In Push-T, this interface supports joint action/state training and direct closed-loop execution through one autoregressive language model.

\newpage

%% file: sec/methods.tex
\section{Method}

We formulate \emph{Spatial Language Modeling} as autoregressive completion over grammar-structured token sequences for Push-T. Scene geometry, the goal, pusher actions, and optional future states are represented by coordinate and semantic tokens in a \emph{Spatial Language Sentence} (SLS). Training sequences contain observed states and rollout targets. At inference, an observed-history prefix prompts the model to sample a completion, which is parsed into pusher actions. The grammar defines the sequence format and parser.

\subsection{Spatial Language Modeling}\label{sec:slm}

\vspace{1mm}
\noindent
\textbf{Tokenization and Vocabularies.}\label{sec:voca}
We define the planar workspace as an axis-aligned rectangle with minimum and maximum corners $\mathbf{b}_{min},\mathbf{b}_{max}\in\mathbb{R}^2$. The workspace is discretized into a regular grid with cell size $v_s$ and dimensions $\mathbf{v_g}=(v_1,v_2)$, where $\mathbf{v_g} = \lceil (\mathbf{b}_{max} - \mathbf{b}_{min})/{v_s} \rceil$. All possible grid locations form the coordinate vocabulary $\mathcal{Q}_{coord}$.

We project scene geometry onto the grid and represent occupied contour cells as coordinate tokens. Multiple contour points in the same cell map to the same token. Each grid coordinate is represented by one model token with a unique integer ID under a fixed row-major indexing convention. The tuple $\mathbf{p}=(p_1,p_2)$ is a readable label for this single token. The grid resolution trades geometric precision against vocabulary size and the number of tokens needed to represent contours; it also quantizes action targets.

The semantic vocabulary $\mathcal{Q}_{semantic}$ identifies state blocks, entity types, and pusher motion commands. The Push-T symbols and their roles are specified in \S~\ref{sec:tasks} and the supplementary material.

\begin{figure}[t]
\centering
\begin{minipage}[t]{0.60\linewidth}
\centering
\vspace{1mm}
\includegraphics[width=\linewidth]{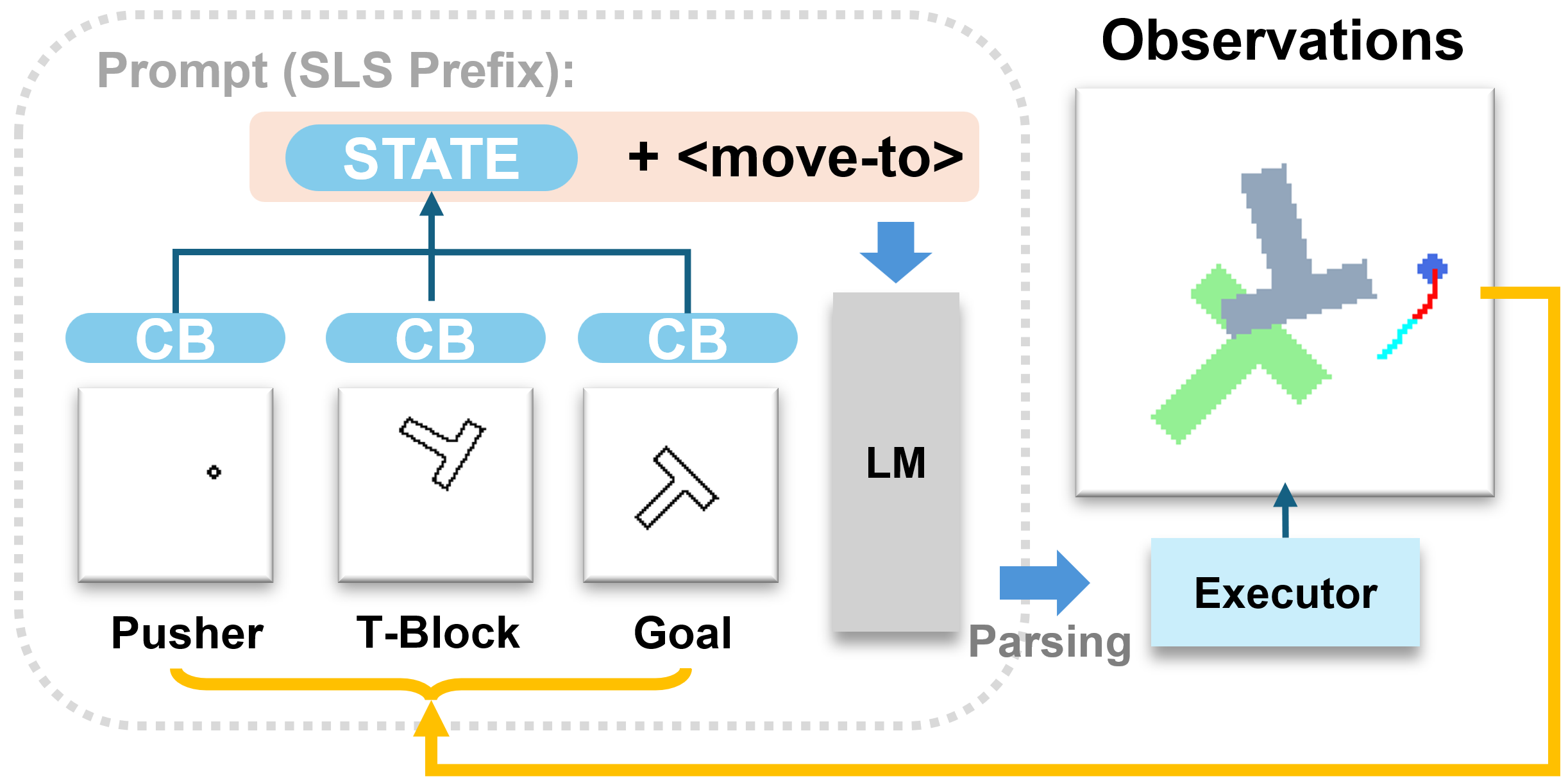}
\caption{Push-T workflow. The model completes an observed-history SLS prefix into actions, and new observations close the control loop.}
\label{fig:pipeline3}
\end{minipage}\hfill
\begin{minipage}[t]{0.37\linewidth}
\centering
\captionof{table}{Push-T sequence format. Each \texttt{COORD} is one token; control decoding omits optional state blocks.}
\label{tab:grammar}
\scriptsize
\begin{tabularx}{\linewidth}{lX}
\toprule
\textbf{Symbol} & \textbf{Definition} \\
\midrule
\multicolumn{2}{l}{\textbf{Context-Free Grammar}} \\
\midrule

$\mathsf{SEQ}$
& $\mathsf{STATE}~\mathsf{MOVE\_TO}^{*}$ \\

$\mathsf{MOVE\_TO}$
& $\texttt{move-to}~\mathsf{CB}~[~\mathsf{STATE}~]$ \\

$\mathsf{STATE}$
& \parbox[t]{\linewidth}{\raggedright\ttfamily state pusher $\mathsf{CB}$\textsuperscript{+} tblock $\mathsf{CB}$\textsuperscript{+} goal $\mathsf{CB}$\textsuperscript{+}} \\

$\mathsf{CB}$
& $\texttt{COORD}$ \\

\midrule
\multicolumn{2}{l}{\textbf{Lexical Tokens}} \\
\midrule

\texttt{COORD}
& One ID in $\mathcal{Q}_{coord}$, displayed as $(p_1,p_2)$ with $0\leq p_j<v_j$ \\

\bottomrule
\end{tabularx}
\end{minipage}
\vspace{-6mm}
\end{figure}

\vspace{1mm}
\noindent
\textbf{Grammar as an Interface.}
An SLS $\mathbf{s}$ is an ordered sequence of tokens from the union of the two vocabularies:
\begin{equation}
    \mathbf{s} \in (\mathcal{Q}_{coord} \cup \mathcal{Q}_{semantic})^n,
\end{equation}
where $n$ is the sequence length. A context-free grammar (CFG; Tab.~\ref{tab:grammar}) specifies the ordering and combination of tokens in training sequences. The $\mathsf{SEQ}$ rule defines the rollout format. A compiler serializes observations and actions into this format, and a parser converts generated action blocks back into target coordinates. For control, a grammar state machine excludes future-state blocks; action coordinates are sampled autoregressively.

\subsection{Application to Push-T}\label{sec:tasks}
We instantiate Spatial Language Modeling for Push-T~\cite{chi2025diffusion}, where a planar pusher moves a T-shaped block to a specified target pose. The input describes the pusher, block, and goal geometry; outputs specify pusher target coordinates and optional subsequent scene states.

We define a coordinate system and a bounding rectangle on the tabletop, then discretize it into coordinate tokens. Tab.~\ref{tab:grammar} summarizes the grammar and token definitions, and Fig.~\ref{fig:pipeline3} illustrates the Push-T workflow.

\noindent
\textbf{Scene Description.} We project the T-block, pusher, and target region onto the plane, keep their outer contours, and represent each entity using the coordinate sequence $\mathsf{CB}$\textsuperscript{+} of its occupied contour cells. These sequences form the CFG nonterminal $\mathsf{STATE}$.

\noindent
\textbf{Action Definition and Future Prediction.} A $\mathsf{MOVE\_TO}$ block starts with the semantic token \texttt{move-to}, followed by one coordinate token specifying the target of the pusher center and, optionally, a $\mathsf{STATE}$ describing the subsequent scene. Observed subsequent states supply scene-transition targets during training. Action generation and state prediction use the same autoregressive model and next-token objective, so both targets update the shared parameters used to generate actions. During control, the model produces consecutive action blocks, reducing the output token count, and closes the loop through new observations.

\noindent
\textbf{Rollout Format.} Each Push-T rollout starts with $\mathsf{STATE}$ followed by a temporally ordered sequence of $\mathsf{MOVE\_TO}$ productions, matching the $\mathsf{SEQ}$ rule. Each step is one pusher movement and corresponds to one scene frame. A $\mathsf{MOVE\_TO}$ block specifies the target coordinate for that step and optionally the resulting $\mathsf{STATE}$. Rollout discretization makes consecutive pusher positions adjacent on the grid, yielding local displacements in the serialized trajectories.

\begin{figure}[t]
    \centering
    \includegraphics[width=0.9\linewidth]{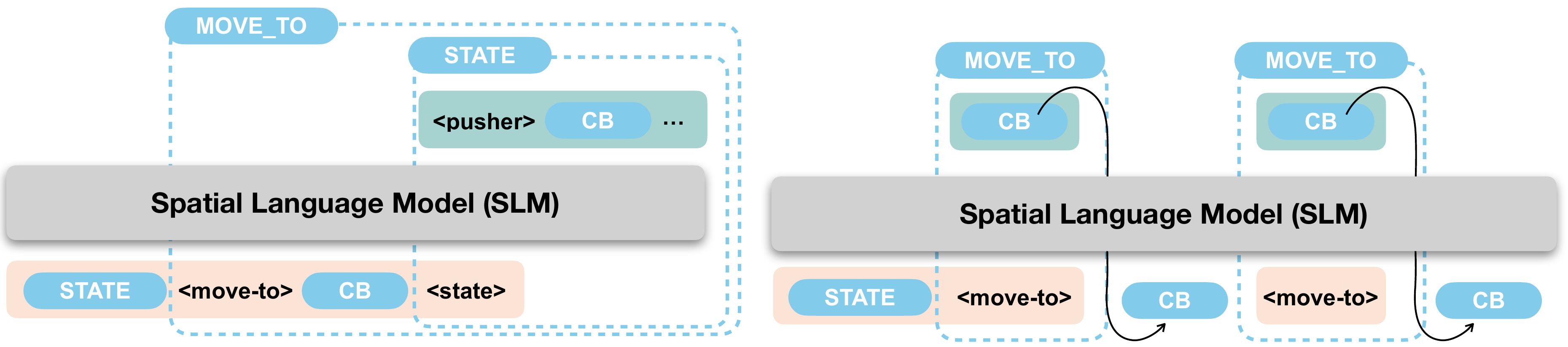}\vspace{-2mm}
    \caption{SLS completion for action-conditioned state prediction (left) and action-only control decoding (right). Generated action blocks are parsed into pusher target coordinates.}
    \label{fig:infer}
      \vspace{-1mm}
\end{figure}

\noindent
\textbf{Training and Inference.}
We compile rollouts into SLS sequences and optimize next-token cross-entropy:
\begin{equation}
    \mathcal{L}(\theta)=-\sum_{i=1}^{n}m_i\log p_{\theta}(s_i\mid s_{<i}),
\end{equation}
where $m_i\in\{0,1\}$ indicates whether position $i$ contributes to the loss. In expert-demonstration training, $m_i=1$ for all sequence positions. During random-play pretraining, $m_i=0$ only for the action-coordinate token immediately following \texttt{move-to}; all remaining tokens, including the command marker and state coordinates, contribute to the loss. Subsequent $\mathsf{STATE}$ blocks are independently retained with probability 30\%; when omitted, their $\mathsf{MOVE\_TO}$ blocks contain only the pusher target. Retained blocks provide scene-transition targets, while omitted blocks create consecutive-action subsequences within the same training format.

For closed-loop control, the prefix comprises three temporally consecutive $\mathsf{STATE}$--$\mathsf{MOVE\_TO}$ segments and ends with \texttt{move-to} for the next action coordinate. A grammar state machine excludes future $\mathsf{STATE}$ blocks while the model samples action coordinates autoregressively (Fig.~\ref{fig:infer}). Parsed targets are executed, and new observations update the history for the next prefix (Fig.~\ref{fig:pipeline3}). The execution schedule is specified in Appendix~\ref{sec:pusht-supp}.

\noindent
\textbf{Pretraining with Random-Play.} We collect rollouts in which the pusher moves randomly and may contact the T-block. Action-coordinate tokens supply the conditioning context, with their loss positions masked. Subsequent state tokens provide scene-transition targets conditioned on these random actions. Task-specific training then supervises demonstration sequences, including action coordinates and retained state blocks. Both stages update the same model parameters, combining transition data from random play with expert action supervision.

%% file: sec/experiment.tex
\section{Experiments}
\label{sec:experiments}
We evaluate Spatial Language Modeling through closed-loop pushing, action-conditioned future-state prediction, and ablations of joint training and random-play pretraining. Simulation and real-robot experiments examine how the shared sequence model generates executable actions and predicts the geometric effects of pushing. Implementation details are provided in Appendix~\ref{sec:pusht-supp} and Appendix~\ref{sec:baselines}.

\subsection{Push-T Setup and Evaluation}
We discretize the workspace into a \(128\times128\) grid and train a 462M-parameter Transformer using the Qwen2 architecture~\cite{team2024qwen2} from scratch. Training takes 5 hours on four NVIDIA A100 GPUs.

\noindent
\textbf{Simulation Data.}
We use the released 200 expert demonstrations from Diffusion Policy~\cite{chi2025diffusion}. Our model is trained for 200 epochs, with one checkpoint evaluated every 5 epochs. Our model receives rasterized contour coordinates, while the reported simulation baselines use state inputs.

\noindent
\textbf{Real-Robot System and Data.}
A stick mounted on the robot arm serves as the pusher, and an industrial camera observes the tabletop from above (Fig.~\ref{fig:teaser}). AprilTags~\cite{olson2011apriltag} on the T-block and tabletop provide localization, from which the known block geometry is converted to contour coordinates in the calibrated tabletop frame. The arm, camera, and tabletop share this coordinate system.

We collect 40 successful expert demonstrations and 20 random-play episodes. The random-play data contains 5,229 frames in total and is used for action-conditioned state-prediction pretraining, with action-coordinate tokens excluded from the loss. Each collected episode contains no more than 300 steps. We compare against the state-based DP-Coord, DP-Tag, and ACT adaptations described in Appendix~\ref{sec:baselines}. These policy baselines are trained on the expert demonstrations, while our full method also uses random-play data.

\noindent
\textbf{Coverage and Success.}
Let \(B_t\) be the set of grid cells occupied by the filled T-block region at step \(t\), and let \(G\) be the corresponding set for the target region. Coverage is
\begin{equation}
    C_t=\frac{|B_t\cap G|}{|G|}.
    \label{eq:coverage}
\end{equation}
A test episode is successful and terminates when \(C_t\geq0.95\). Otherwise, evaluation ends after 300 steps (frames), each corresponding to one pusher movement. Coverage is computed from filled regions on the discrete grid, while model inputs represent their contours.

In the real-robot evaluation, each method is tested from 20 initial states, with starting poses manually matched across methods as closely as possible. Our model uses a three-frame observation-history window. The implemented DP baselines use two observation frames and ACT uses one; their input encodings and action-chunk execution settings are specified in Appendix~\ref{sec:baselines}.

For real-robot results, \emph{Max} is the mean of the maximum coverage reached across the test episodes:
\begin{equation}
    \mathrm{Max}=\frac{1}{N}\sum_{e=1}^{N}\max_{t\in\mathcal{T}_e} C_{e,t},
    \label{eq:real_max}
\end{equation}
where $N$ is the number of test episodes and $\mathcal{T}_e$ contains the evaluated steps of episode $e$. Success rate is the fraction of these episodes that reach the coverage threshold within 300 steps.

\begin{figure}[t]
\centering
\begin{minipage}[t]{0.65\linewidth}
\centering
\captionof{table}{Push-T results. In simulation (left), \emph{Max} is the highest average coverage across evaluated checkpoints and \emph{Avg} is the average coverage over the last 10 checkpoints. Real-robot results (right) report success rate (\emph{SR}) and the mean of episode-level maximum coverage (\emph{Max}; Eq.~\ref{eq:real_max}) over 20 episodes per method. \emph{Action-only} deletes future-state blocks and omits random-play pretraining; \emph{w/o P.} omits pretraining but retains joint action/state training.}\label{tab:pusht_comparison}

\begin{minipage}[t]{0.48\linewidth}
\centering
\begin{tabular}{l|cc}
\hline
Methods     & Max & Avg \\ \hline
LG~\cite{mandlekar2021matters} & 0.67 & 0.61 \\
IBC~\cite{florence2022implicit} & 0.90 & 0.84 \\
BET~\cite{shafiullah2022behavior} & 0.79 & 0.70 \\
DP-C~\cite{chi2025diffusion} & \textbf{0.95} & 0.91 \\
DP-T~\cite{chi2025diffusion} & \textbf{0.95} & 0.79 \\
Ours & \textbf{0.95} & \textbf{0.93} \\ \hline
\end{tabular}
\end{minipage}
\hfil 
\begin{minipage}[t]{0.45\linewidth}
\centering
\setlength{\tabcolsep}{3.5pt}

\begin{tabular}{l|cc}
\hline
Methods & SR & Max \\
\hline
DP-Coord & 0.45 & 0.89 \\
DP-Tag & 0.50 & 0.89 \\
ACT & 0.65 & 0.86 \\
Ours & \textbf{0.80} & \textbf{0.95} \\
\hline
Action-only & 0.05 & 0.67 \\
Ours w/o P. & 0.65 & 0.92 \\
\hline
\end{tabular}
\end{minipage}\hfill
\end{minipage}\hfill
\begin{minipage}[t]{0.32\linewidth}
\centering
\vspace{5mm}
\includegraphics[width=\linewidth]{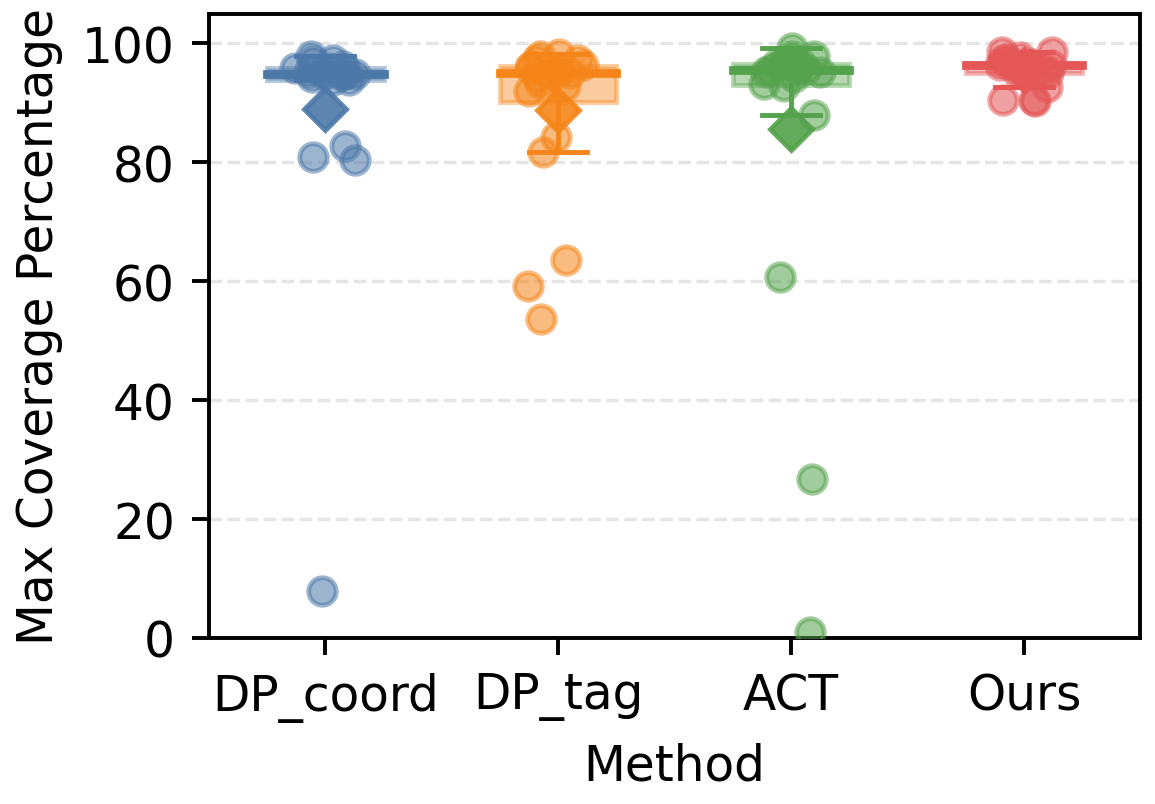} 
\captionof{figure}{Distribution of episode-level maximum coverage in the real-robot Push-T evaluation.}\label{fig:mean_max_coverage_box}
\end{minipage}
\end{figure}

\begin{figure}[t]
    \centering
    \includegraphics[width=\linewidth]{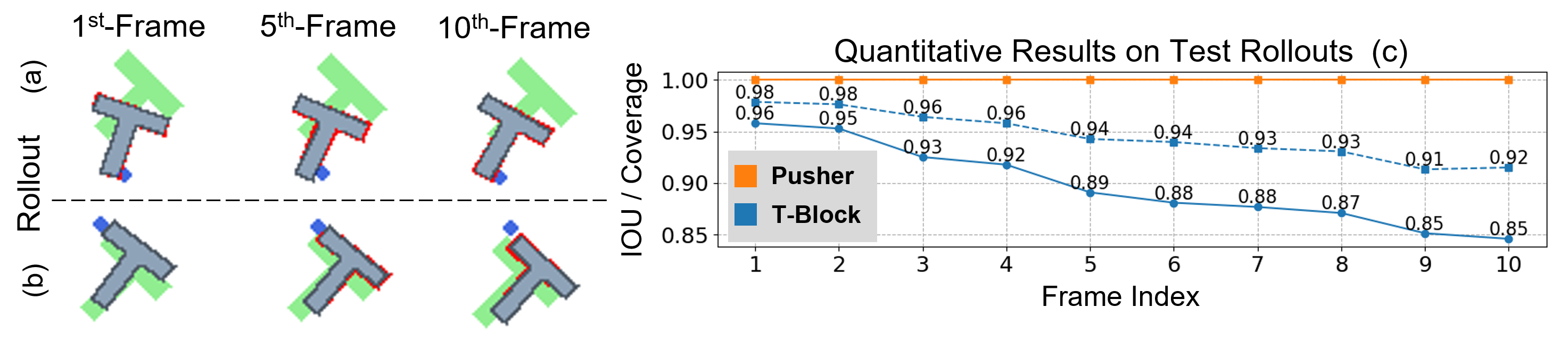}\vspace{-1mm}
    \caption{Future-state prediction on Push-T. Given a manually annotated action trajectory, the model predicts successive states using its own previous predictions as context. Reference states are obtained by executing the same actions in the simulator.
Subfigures (a) and (b) show two example rollouts, where red pixels indicate the prediction-error regions of the T-block on the rasterized workspace.
Subfigure (c) reports the average state prediction accuracy at each successive step (frame) over the test rollouts, where dashed lines denote coverage and solid lines denote IoU.
}
    \label{fig:exp_pusht}
      \vspace{-4mm}
\end{figure}

\begin{figure}[t]
    \centering
    \includegraphics[width=\linewidth]{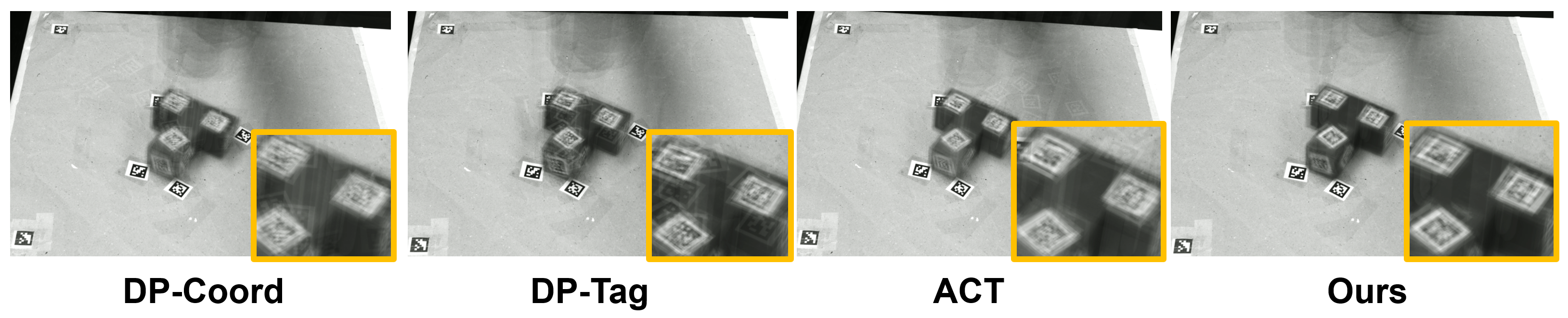}
    \caption{For each method, images are averaged over the test episodes at their respective maximum-coverage frames. This visualization summarizes alignment across episodes.}
    \label{fig:pusht_visual}
      \vspace{-6mm}
\end{figure}

\subsection{Push-T Results}
\noindent
\textbf{Simulation.}
Table~\ref{tab:pusht_comparison} (left) reports the highest mean coverage across checkpoints (\emph{Max}) and the mean coverage over the last ten checkpoints (\emph{Avg}). Our model matches the best Max of 0.95 achieved by DP-C and DP-T, while reaching the highest Avg of 0.93, compared with 0.91 and 0.79, respectively. Its average over the last ten checkpoints is close to its best coverage, showing that strong performance extends beyond the single best checkpoint. These results demonstrate that language-model completion of spatial sequences can produce an effective closed-loop pushing policy.

\noindent
\textbf{Future-State Prediction.}
Figure~\ref{fig:exp_pusht} shows the model predicting T-block translation and rotation along manually annotated action trajectories. Each predicted state supplies the context for the next prediction, and the simulator provides reference states under the same actions. The T-block's coverage changes from 0.98 at the first frame to 0.92 at the tenth, while IoU changes from 0.96 to 0.85. Although error accumulates over the rollout, substantial geometric overlap remains after ten steps (frames). Together, the examples and frame-wise metrics show that the same spatial language model used for action generation also learns action-conditioned scene transitions.

\noindent
\textbf{Real Robot.}
Our full method achieves a success rate of 0.80, compared with 0.45 for DP-Coord, 0.50 for DP-Tag, and 0.65 for ACT (Tab.~\ref{tab:pusht_comparison}, right). It also reaches the highest mean episode-level maximum coverage of 0.95, compared with 0.89 for both DP variants and 0.86 for ACT. The language model's generated coordinates therefore support successful physical execution as well as close geometric alignment with the target.

Figure~\ref{fig:mean_max_coverage_box} further shows that our episode-level peak coverage is concentrated around 90\%--100\%. The baselines include low-overlap episodes, while the full model brings the block close to the target across the tested initial states. Figure~\ref{fig:pusht_visual} provides a complementary view: averaging the maximum-coverage frames yields a clearer T-block outline and fewer ghosting artifacts for our method, indicating more concentrated block positions and orientations. The numerical and visual results together show improvements in task completion and target alignment.

\noindent
\textbf{Ablations.}
We compare two training variants with the full method. \emph{Action-only} deletes the future $\mathsf{STATE}$ blocks from the demonstration sequences and uses no random-play pretraining. \emph{Ours w/o P.} retains joint action and future-state training on the demonstrations, also without pretraining. Adding future-state blocks raises success from 0.05 to 0.65 and mean episode-level maximum coverage from 0.67 to 0.92. Joint training thus improves both task completion and geometric alignment. It gives the shared model targets for the actions to take and the scene changes that follow, making future-state prediction a useful component of policy training.

Random-play pretraining further raises success from 0.65 to 0.80 and maximum coverage from 0.92 to 0.95. This stage supplies additional state-transition supervision conditioned on recorded pusher actions, followed by demonstration training for goal-directed control. The two comparisons show complementary benefits from joint action/state training on demonstrations and transition pretraining on random interactions. Both stages use the common spatial vocabulary and update the same autoregressive model.

\noindent
\textbf{Inference Throughput.}
On a single A100 GPU, language-model throughput is approximately 7000 tokens/s during prefilling and 800 tokens/s during decoding. During control, the grammar state machine selects the action-only sequence format, in which each target requires one \texttt{move-to} token and one coordinate token. Generating 100 action targets therefore requires about 200 output tokens, corresponding to about 0.25~s of decoding at the measured throughput. The 20-step execution chunk requires about 40 tokens, or approximately 0.05~s of decoding. This compact output format supports rapid action-chunk generation while the same model is trained jointly on actions and future states.

%% file: sec/Conclusion.tex
\section{Limitations}
\label{sec:limitation}
Our method has several limitations. First, the real-robot system relies on calibrated AprilTag localization and known object geometry; end-to-end learning from raw sensor observations has not been evaluated. Second, the grammar is manually designed, while grid resolution trades geometric and action precision against vocabulary size and sequence length. Adapting to new entities, action types, or task structures may require changes to the representation. Third, evaluation focuses on planar Push-T, so generalization to other object geometries, manipulation tasks, and three-dimensional scenes remains to be established. Finally, errors accumulate during autoregressive state rollouts, and prediction over longer horizons requires further evaluation. This limitation concerns state prediction; closed-loop control updates its history with newly observed states.

\section{Conclusion}
We presented Spatial Language Modeling, which expresses scene geometry, goals, action targets, and future states as grammar-structured coordinate sequences. One autoregressive model learns action generation and action-conditioned state prediction through a shared next-token objective. Training combines transition pretraining on random interactions with joint action and state supervision on expert demonstrations. During closed-loop execution, the model generates action targets directly and updates its context using new observations. Experiments demonstrate effective control in simulated and real-world Push-T, while rollouts conditioned on supplied actions capture successive geometric scene changes. Training comparisons show benefits from interleaved action and state sequences, with additional gains from random-play pretraining. These findings support shared spatial modeling as a practical way to connect action-conditioned transition supervision with goal-directed policy learning.

%% file: sec/supplementary.tex
\section{Model Implementation}
We employ a 462M-parameter Qwen2 model~\cite{team2024qwen2} and train it from scratch. Specifically, the model consists of 7 transformer layers with a hidden dimension of 1536 and 24 attention heads. Rotary Position Embeddings (RoPE) are used for positional encoding. Push-T uses 16,384 coordinate tokens for its $128\times128$ grid, together with the semantic tokens defining its sequence format. We pack multiple spatial language sentences into training samples with a maximum context length of 13,000.

\section{Push-T Representation and Execution Details}
\label{sec:pusht-supp}
We provide the grammar definition for the Push-T task~\cite{chi2025diffusion}. In Push-T, a pusher moves a T-shaped block to a goal region. The observation and action are both represented on a discretized 2D grid and serialized as a Spatial Language Sentence (SLS).

\vspace{2mm}
\noindent
\textbf{2D Representation and Vocabularies.}
We discretize the workspace in 2D. Each grid location is encoded as one coordinate token $\mathsf{CB}\in\mathcal{Q}_{coord}$. In the simulator, we render the pusher, the T-block, and the goal region, extract their edge pixels, and represent each entity by its contour coordinate tokens.

\vspace{2mm}
\noindent
\textbf{State (Scene) Description.}
A Push-T state is composed of three components: the pusher geometry, the T-block geometry, and the goal-region geometry. We define the nonterminal $\mathsf{STATE}$ to serialize these components as
\[
\mathsf{STATE} \rightarrow \texttt{state}~\texttt{pusher}~\mathsf{CB}^{+}~\texttt{tblock}~\mathsf{CB}^{+}~\texttt{goal}~\mathsf{CB}^{+}.
\]
Here, the terminal tokens \texttt{state}, \texttt{pusher}, \texttt{tblock}, and \texttt{goal} belong to $\mathcal{Q}_{semantic}$, and each $\mathsf{CB}^{+}$ is a (sorted) sequence of coordinate tokens that describes the corresponding entity.

\vspace{2mm}
\noindent
\textbf{Action  Description.}
We represent a pushing action as moving the pusher to a target 2D coordinate. Each such movement is one step and corresponds to one scene frame. Accordingly, the nonterminal $\mathsf{MOVE\_TO}$ is defined as
\[
\mathsf{MOVE\_TO} \rightarrow \texttt{move-to}~\mathsf{CB}~[~\mathsf{STATE}~].
\]
The optional bracket $[~\mathsf{STATE}~]$ indicates that we may append the next observation state after executing the action, enabling multiple actions to be represented within a single sequence.

The Push-T task is then defined by the 2D production rule:
\[
\mathsf{SEQ}_{2D} \rightarrow \mathsf{STATE}~\mathsf{MOVE\_TO}^{*}.
\]

The sequence format consists of an initial $\mathsf{STATE}$ and temporally ordered action blocks, optionally interleaved with resulting states. Training trajectories follow this format. During control, a grammar state machine excludes future $\mathsf{STATE}$ blocks; action coordinates are sampled autoregressively and parsed into pusher targets.

\vspace{1mm}
\noindent
\textbf{Closed-loop execution at test time.}
At test time, the prefix contains three temporally consecutive $\mathsf{STATE}$--$\mathsf{MOVE\_TO}$ segments and ends with \texttt{move-to} for the next target coordinate. We execute the first 20 predicted \texttt{move-to} targets from the action-only continuation, one per step (frame), then update the history with new observations for the next prompt.

\vspace{1mm}
\noindent
\textbf{Training objectives.}
Task demonstration training uses next-token prediction over the serialized trajectories. Random-play pretraining masks only the target-coordinate token immediately following each \texttt{move-to} marker; the marker and all other tokens remain supervised. Target coordinates remain in the context to condition subsequent state predictions. Action generation and state prediction share model parameters.

\vspace{1mm}
\noindent
\textbf{Success criterion.}
Let $B_t$ be the set of filled grid cells occupied by the T-block at time $t$, and let $G$ be the set of filled grid cells in the target region. Coverage is
\[
\operatorname{Coverage}(t)=\frac{|B_t\cap G|}{|G|}.
\]
Evaluation episodes terminate successfully at $0.95$ coverage or unsuccessfully after 300 steps (frames), corresponding to at most 300 pusher movements. Coverage uses filled grid regions; model inputs use their contours.

For the real-robot results, \emph{Max} averages each episode's maximum coverage over the 20 test episodes, as defined in Eq.~\ref{eq:real_max}. \emph{SR} is the fraction of successful episodes. The simulation table instead reports the highest mean coverage across evaluated checkpoints and the mean over the last 10 checkpoints.

\section{Full Symbol and Token Definitions}
This section complements the brief grammar description in the main paper by explaining the symbols used for Push-T.

\vspace{2mm}
\noindent
\textbf{Vocabularies.}
All SLS tokens come from the union of the coordinate vocabulary $\mathcal{Q}_{coord}$ and the semantic vocabulary $\mathcal{Q}_{semantic}$. Coordinate tokens represent discretized locations on the planar grid.

The Push-T semantic tokens are:
\[
\mathcal{Q}_{semantic} = \{\texttt{state},\ \texttt{pusher},\ \texttt{tblock},\ \texttt{goal},\ \texttt{move-to}\}.
\]

\vspace{2mm}
\noindent
\textbf{Lexical token: \texttt{COORD}.}
Each \texttt{COORD} is one vocabulary token identifying one discrete grid cell. The tuple $(x,y)$ is readable notation for that single token ID.

\vspace{2mm}
\noindent
\textbf{Coordinate block symbol: $\mathsf{CB}$.}
In the CFG, a coordinate block $\mathsf{CB}$ expands to a single \texttt{COORD} token. A repetition $\mathsf{CB}^{+}$ denotes a non-empty list of coordinates representing the discretized geometry of an entity.

\vspace{2mm}
\noindent
\textbf{2D task symbols (Push-T).}
The 2D production $\mathsf{SEQ}_{2D}$ defines a Push-T rollout:
\[
\mathsf{SEQ}_{2D} \rightarrow \mathsf{STATE}~\mathsf{MOVE\_TO}^{*}.
\]

\vspace{1mm}
\noindent
\emph{State block:} $\mathsf{STATE}$.
$\mathsf{STATE} \rightarrow \texttt{state}~\texttt{pusher}~\mathsf{CB}^{+}~\texttt{tblock}~\mathsf{CB}^{+}~\texttt{goal}~\mathsf{CB}^{+}$ describes one observation, where the coordinate sequences correspond to edge pixels extracted from the rendered pusher, T-block, and goal region.

\vspace{1mm}
\noindent
\emph{Action block:} $\mathsf{MOVE\_TO}$.
$\mathsf{MOVE\_TO} \rightarrow \texttt{move-to}~\mathsf{CB}~[~\mathsf{STATE}~]$ represents one action that moves the pusher to the target coordinate $\mathsf{CB}$. The optional $[~\mathsf{STATE}~]$ allows appending the next observation after execution.

\vspace{2mm}
\noindent
\textbf{Canonicalization (coordinate serialization).}
We sort each entity-specific geometry list $\mathsf{CB}^{+}$ by a fixed grid order ($x$, then $y$). Sorting is confined to each entity block.

\input{sec/supp_baselines}

%% file: sec/supp_baselines.tex
\providecommand{\cusR}{\mathbb{R}}

\subsection{Baseline Implementation Details}
\label{sec:baselines}

All three baselines (DP-Coord, DP-Tag, and ACT) operate in the state-based
real-world Push-T setting. Their per-frame inputs describe the manipulated
T-block together with the pusher-center coordinate
$\mathbf{p}_t\in\cusR^{2}$. Our SLS input serializes contour coordinates for
the pusher, T-block, and goal (\S\ref{sec:tasks}). The baselines use two T-block
representations:
\begin{itemize}
  \item \textbf{Object contour coordinates} (used by DP-Coord and ACT): the
    T-block contour coordinates $X_t=\{\mathbf{x}_{t,i}\}_{i=1}^{K}$,
    $\mathbf{x}_{t,i}\in\cusR^{2}$, taken from the discretized outer contour
    used in our $\mathsf{STATE}$ block. The number of occupied cells varies
    per frame, so each frame is sentinel-padded to a dataset-wide fixed count
    $K{=}100$ with $\bar{\mathbf{x}}=(-1,-1)$ marking empty slots.
  \item \textbf{AprilTag corner coordinates} (used by DP-Tag): the
    $S=4\,\lvert\mathcal{T}\rvert=12$ corner coordinates from the
    $\lvert\mathcal{T}\rvert=3$ AprilTags on the top surface of the T-block
    (four corners each), placed in a fixed slot order with no padding.
\end{itemize}
For all baselines, the action is the absolute target pusher coordinate
$\mathbf{a}_t\in\cusR^{2}$, i.e.\ the same quantity our $\mathsf{MOVE\_TO}$
production emits. Relative to their default implementations, the observation
encoding (and, for ACT, the token construction) is modified; the action
heads and training objectives are inherited unchanged. Diffusion Policy uses
two observation frames, ACT uses one frame, and our model uses three frames.

\paragraph{DP-Coord / DP-Tag (Diffusion Policy).}
Both Diffusion Policy baselines use low-dimensional Diffusion
Policy~\cite{chi2025diffusion}: a conditional
$1$-D U-Net $\boldsymbol{\epsilon}_\theta$ trained with DDPM and conditioned on the
observation as a \emph{global conditioning} vector. We adapt the original
Push-T low-dimensional input, which contains nine ground-truth keypoints,
to a flattened geometry vector concatenated with the pusher coordinate,
\begin{equation}
  \mathbf{o}_t \;=\;
  \bigl[\,\operatorname{vec}(X_t)\ \big\Vert\ \mathbf{p}_t\,\bigr]
  \;\in\;
  \begin{cases}
    \cusR^{2K+2}, & \text{DP-Coord (contour coords)},\\[2pt]
    \cusR^{2S+2}, & \text{DP-Tag (tag coords)},
  \end{cases}
\end{equation}
where $\operatorname{vec}(\cdot)$ stacks the coordinates in their fixed order.
Each dimension is normalized to $[-1,1]$ (limits normalizer). DP-Coord feeds
the sentinel-padded contour through the fixed-width global-conditioning
path with no explicit padding mask. Stacking
$n_o{=}2$ observation steps yields the conditioning
$\mathbf{c}_t=[\mathbf{o}_{t-1};\mathbf{o}_t]$ passed to
$\boldsymbol{\epsilon}_\theta$. The output is the default Diffusion Policy
receding-horizon scheme: denoise an action trajectory
$\hat{\mathbf{a}}_{t:t+H}\in\cusR^{H\times 2}$ of horizon $H{=}16$ and execute
the first $n_a{=}8$ steps before re-planning.

\paragraph{ACT.}
The default ACT~\cite{zhao2023learning} (DETR-VAE) feeds a ResNet image
feature map ($H'\!\times\!W'$ tokens) as the cross-attention memory of its
transformer decoder, plus a proprioception token from the agent state and a
CVAE latent token. For our state-based task, we replace the CNN feature
tokens with \emph{per-coordinate} tokens over the object contour
coordinates. Each coordinate is lifted to the hidden dimension $d$ by a single
shared learnable linear map $\phi:\cusR^{2}\!\to\!\cusR^{d}$ and combined with
a learnable per-slot positional embedding $\mathbf{e}_i$,
\begin{equation}
  \mathbf{m}_{t,i} \;=\; \phi(\mathbf{x}_{t,i}) + \mathbf{e}_i,
  \qquad i=1,\dots,K,
\end{equation}
giving $K$ memory tokens. The pusher coordinate is encoded by the \emph{same}
kind of learnable linear that default ACT applies to the agent state,
$\mathbf{m}^{\mathrm{prop}}_t=\psi(\mathbf{p}_t)$,
$\psi:\cusR^{2}\!\to\!\cusR^{d}$, and prepended together with the CVAE latent
token. Padded contour slots are excluded from attention through a key-padding
mask $\mu_{t,i}=\mathbf{1}[\mathbf{x}_{t,i}=\bar{\mathbf{x}}]$, so a single
model serves the variable-length contour. The CVAE encoder, the latent prior,
and the $L_1$ + KL training objective are identical to default ACT. The
decoder uses $H{=}16$ learnable query tokens and a linear action head to emit
the chunk $\hat{\mathbf{a}}_{1:H}\in\cusR^{H\times 2}$; at inference the latent
is fixed to its prior mean ($\mathbf{0}$). ACT conditions on a single
observation step ($n_o{=}1$) and executes the first $n_a{=}8$ of the $16$
predicted actions per re-plan.